# Architectural Insights into Knowledge Distillation for Object Detection: A Comprehensive Review

Mahdi Golizadeh, Nassibeh Golizadeh, Mohammad Ali Keyvanrad, and Hossein Shirazi

***Abstract*—** Object detection has achieved remarkable accuracy through deep learning, yet these improvements often come with increased computational cost, limiting deployment on resource-constrained devices. Knowledge Distillation (KD) provides an effective solution by enabling compact student models to learn from larger teacher models. However, adapting KD to object detection poses unique challenges due to its dual objectives—classification and localization—as well as foreground-background imbalance and multi-scale feature representation. This review introduces a novel architecture-centric taxonomy for KD methods, distinguishing between CNN-based detectors (covering backbone-level, neck-level, head-level, and RPN/RoI-level distillation) and Transformer-based detectors (including query-level, feature-level, and logit-level distillation). We further evaluate representative methods using the MS COCO and PASCAL VOC datasets with mAP@0.5 as performance metric, providing a comparative analysis of their effectiveness. The proposed taxonomy and analysis aim to clarify the evolving landscape of KD in object detection, highlight current challenges, and guide future research toward efficient and scalable detection systems.

## I. INTRODUCTION

OBJECT Detection (OD) is a fundamental task in computer vision with applications ranging from autonomous driving and surveillance to robotics and medical diagnostics. Modern detectors—such as Faster R-CNN [1], YOLO [2], SSD [3], and Transformer-based models like DETR [4]—have achieved impressive accuracy, but at the cost of increased model complexity and computation. This creates practical challenges for deployment on resource-constrained devices like drones, mobile phones, and embedded platforms.

Knowledge Distillation (KD) [5, 6] offers a promising solution by enabling compact student models to learn from larger, high-performing teacher models without major changes to the architecture. However, adapting KD to OD is non-trivial due to the dual nature of the task—requiring both classification and localization—as well as issues such as foreground-background imbalance and multi-scale processing.

While prior surveys [7, 8] have organized KD methods by task type or knowledge signal, they fall short in connecting distillation strategies to the internal architecture of modern OD models. This architectural dimension is crucial, as OD pipelines include modular components backbones, necks, detection heads modules that behave differently under distillation, for example, backbone distillation often focuses on aligning deep semantic features across the full image, while head-level distillation must handle task-specific outputs like bounding box coordinates, which require careful treatment of spatial precision and uncertainty.

To address this gap, we propose a novel architecture-aware taxonomy of KD methods in object detection. This perspective categorizes existing methods by the structural component they target—backbone, neck, head, and RPN/RoI level for CNN-based and query, feature, and logit level for transformer-based—offering a unified lens for understanding, comparing, and designing distillation strategies.

Our contributions are as follows:

• We introduce a unified taxonomy categorizing Knowledge Distillation (KD) methods based on their architectural targets—specifically, the backbone, neck, head, and RPN/RoI level for CNN-based detectors, and the query, feature, and logit level for Transformer-based detectors—offering a consistent framework across diverse object detection architectures.

• Under this novel architectural lens, we critically analyze representative KD strategies, elucidating the unique challenges encountered when distilling knowledge at different structural levels and uncovering the underlying principles of effective component-specific distillation mechanisms.

• We evaluate these methods across standard benchmarks (COCO and PASCAL VOC) to provide empirical insights and practical guidance.

This paragraph of the first footnote will contain the date on which you submitted your paper for review, which is populated by IEEE.

This work received no specific grant from any funding agency in the public, commercial, or not-for-profit sectors.

(Corresponding author: Mohammad Ali Keyvanrad)

Mahdi Golizadeh is with the Faculty of Electrical and Computer Engineering, University of Tabriz, Tabriz, Iran (e-mail: mahdi.golizadeh@gmail.com).

Nassibeh Golizadeh is with the Faculty of Computer Engineering, Sharif University of Technology, Tehran, Iran (e-mail: nassibeh.golizadeh@ce.sharif.edu).

Mohammad Ali Keyvanrad is with the Faculty of Electrical and Computer Engineering, Malek-Ashtar University of Technology, Tehran, Iran (e-mail: keyvanrad@mut.ac.ir).

Hossein Shirazi is with the Faculty of Electrical and Computer Engineering, Malek-Ashtar University of Technology, Tehran, Iran (e-mail: Shirazi@mut.ac.ir).

## II. BACKGROUND

Object detection has progressed from traditional techniques like Viola-Jones [9] and HOG-SVM [10] to deep neural networks, including CNN-based models such as Faster R-CNN [1], YOLO [2], and SSD [3]. Transformer-based architectures like DETR [4] have further advanced the field by introducing global attention and set-based predictions. These models, while accurate, are computationally intensive and ill-suited for edge deployment.

To improve deployment efficiency, model compression methods such as pruning [11], quantization [12], low-rank factorization [13], and neural architecture search (NAS) [14] have been explored. Among these, Knowledge Distillation (KD) stands out for preserving model structure while increasing precision. Originally proposed by [5] and popularized by [6], KD allows a student network to learn from a teacher's outputs or intermediate representations.

KD was initially successful in image classification, but early attempts to apply it to OD—such as FitNet [15] and Mimic[16]—faced challenges due to the multi-task nature of detection, particularly the need to distill both classification and localization signals simultaneously, and the difficulty in aligning multi-scale features that are common in detection pipelines. OD involves classification and localization, requires multi-scale features, and suffers from foreground-background imbalance, all of which complicate naive KD application. These limitations led to the emergence of task-specific and architecture-sensitive distillation techniques.

Over time, KD methods in OD have expanded into several categories:

• Response-based KD: Distilling classification logits or bounding box scores [17, 18].
• Feature-based KD: Transferring intermediate feature maps with structural or attention-based alignment [19-21].
• Relational KD: Modeling inter-instance dependencies using graphs or contrastive objectives [22, 23].
• Distributional KD: Representing bounding box regression as probability distributions [24].

These modalities reflect different design philosophies and are often applied at different architectural stages of the model. While these categories describe the type of knowledge transferred, existing surveys that attempt to organize the field exhibit limitations—often lacking architectural specificity or failing to connect distillation techniques to model internals.

While several reviews exist, they exhibit key limitations:

• Task-oriented taxonomy [7]: Categorizes KD methods by detection scenario (e.g., incremental OD, remote sensing). This lacks architectural specificity.
• Signal-based taxonomy [8]: Organizes methods based on architectural similarity—homogeneous (teacher and student share architecture) vs. heterogeneous (architecture differs, e.g., Transformer-to-CNN, single stage to two stage). While useful, this view focuses more on signal type and compatibility, offering limited insight into where within the detection pipeline the knowledge is transferred.

Neither survey fully accounts for how knowledge flows through OD architectures—or how specific components (e.g., the neck in FPNs [25] or queries in Transformers) respond differently to distillation.

Modern object detectors are modular by design. CNN-based models include backbones, necks, heads, and RPN/RoI modules, while Transformer-based detectors use encoder-decoder structures, object queries, and logits. Each component presents distinct challenges for effective knowledge transfer so an architecture-centric taxonomy:

• Provides fine-grained insights into component-level distillation challenges.
• Enables clearer analysis of emerging methods aligned with evolving model designs.

Our review builds on this architectural lens to classify and evaluate KD methods in a unified, component-aware manner; for the first time, this also encompasses a dedicated and systematic review of KD methods specifically for Transformer-based OD models within this architectural framework. This framing offers clarity for researchers seeking both theoretical insights and practical guidance on KD in OD.

## III. TAXONOMY OF KD METHODS IN OBJECT DETECTION

This section delves into a structured taxonomy of Knowledge Distillation (KD) techniques as applied to Object Detection (OD). To provide a clear and comprehensive overview, we categorize existing KD methodologies based on the foundational architecture of the object detection models. Our proposed taxonomy is primarily divided into two major streams: KD applied to Convolutional Neural Network (CNN)-based object detectors and KD tailored for Transformer-based object detection frameworks. Within the CNN-based approaches, we will explore distillation strategies at various levels of the architecture, including the backbone, neck, detection head, and the RPN/RoI level (Fig. 1.a). Subsequently, for the emerging Transformer-based detectors, we will examine KD techniques focusing on the query level, as well as the feature (encoder + decoder) and logit components (Fig. 1.b). This classification aims to systematically organize the diverse landscape of KD in OD, highlighting the distinct challenges and opportunities presented by different architectural paradigms.

Many modern knowledge distillation methods for object detection leverage multiple architectural components simultaneously—e.g., combining query, feature, and logit-level supervision in Transformer-based detectors, or distilling across both neck and head in CNNs. In this taxonomy, we categorize each method based on the most influential or novel architectural level it targets, rather than exhaustively listing all applied distillation stages. The primary level is determined using a combination of factors: the authors stated focus or claimed innovation, and supporting evidence from reported ablation studies. This approach ensures consistency while foregrounding the structural insight each method offers into architectural-level distillation.

A visual overview of our taxonomy is presented in Fig. 1, which illustrates the distinction between CNN-based and Transformer-based detectors and the architectural levels at which distillation is applied.

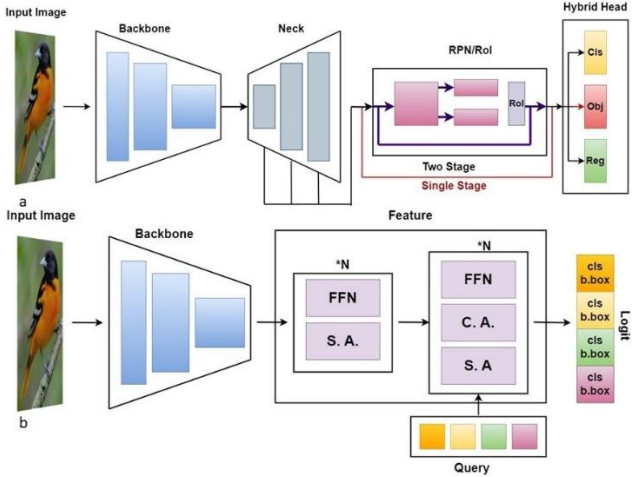

**Fig. 1.** Proposed architecture-centric taxonomy of knowledge distillation (KD) methods in object detection, distinguishing CNN-based (a) and Transformer-based (b) detectors by the architectural level targeted.

*A. CNN-Based Detectors*

1) **Backbone-Level Distillation**

Backbone-level knowledge distillation represents a foundational pillar of KD strategies in CNN-based object detection. These methods aim to transfer semantic and structural knowledge from the teacher's deep feature hierarchy to the student's backbone, which is typically shallower or more efficient. At its simplest, this can involve aligning intermediate feature maps via direct loss functions such as L2 or cosine similarity. However, this naïve feature imitation often fails to capture richer distributional characteristics and spatial context, especially in dense prediction settings. To address these limitations, following approaches have introduced a variety of enhancements grounded in distributional modeling, attention mechanisms, and hierarchical supervision.

A prominent subclass involves adversarial distillation techniques, where a discriminator network is used to distinguish between feature maps generated by the teacher and those from the student for example in [26] the student is trained to produce features that fool the discriminator, thereby implicitly learning the distribution of the teacher's representations (Fig. 2.a). These methods provide stronger supervision than pointwise losses by focusing on the overall distributional alignment. However, they often neglect the foreground-background imbalance in detection, as all pixels contribute equally to the adversarial loss, regardless of semantic relevance. This is particularly problematic for backbone distillation, where dense, low-level features from early layers contain a mix of object-relevant and background information. Without foreground-aware weighting, the student may overfit to background regions, diluting its ability to learn object-centric semantics that are critical for downstream localization and classification.

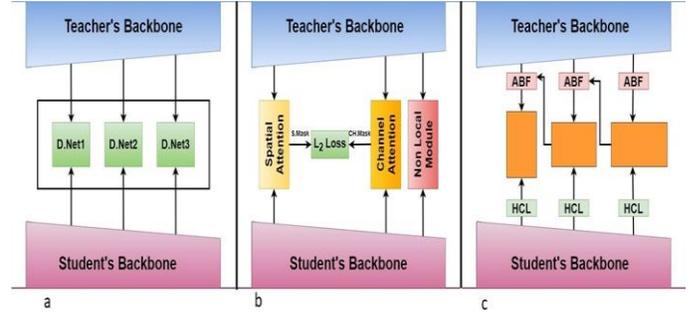

**Fig. 2.** Illustration of three backbone level knowledge distillation strategies: (a) adversarial distillation with discriminator networks; (b) attention-guided and non-local distillation; (c) hierarchical multi-layer distillation with fused feature supervision.

To address this, a separate group of methods employs attention-based methods like attention-guided and non-local distillation strategies in [27]. Here, spatial and channel-wise attention maps derived from the teacher highlight important regions typically object-centric and modulate the feature imitation process. Simultaneously, non-local modules are used to encode long-range dependencies within the feature maps, capturing contextual relationships between different image regions (Fig. 2.b). The student is trained not only to mimic the raw features but also to align with these relational and saliency-aware signals, improving both semantic fidelity and structural representation.

Another emerging line of work emphasizes hierarchical and multi-layer supervision. Unlike traditional layer-to-layer feature matching, these methods fuse information from multiple stages of the teacher's backbone to guide the student. A representative example [19] introduces a residual fusion module that aggregates teacher features across layers using attention-based weighting. The fused representation is then compared to a similarly aggregated student output using a multi-level distillation loss, often computed via spatial pyramid pooling to capture both coarse and fine-grained structure (Fig. 2.c). This approach allows the student to benefit from the teacher's deep representational richness and mitigates the information bottleneck imposed by one-to-one layer matching.

These methods form a coherent taxonomy of backbone-level KD strategies. They vary in how they address distributional complexity, spatial structure, and semantic emphasis, all of which are critical for effective knowledge distillation in object detection.

Table I summarizes key methods that apply backbone-level distillation, highlighting their modifications to the backbone architecture and the specific strategies used for knowledge transfer. Methods vary in their emphasis—ranging from GAN-based imitation, attention modeling, and non-local relational features to hierarchical fusion techniques. This comparative overview helps illustrate the evolution of distillation strategies, especially regarding how they address foreground emphasis, spatial structure, and multi-level feature integration.

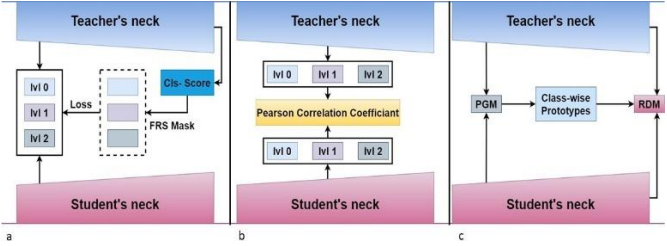

**Fig. 3.** Illustration of three neck-level knowledge distillation approaches, highlighting (a) FRS score, (b) PKD, and (c) class-wise prototype-guided strategies.

### 2) Neck-Level Distillation

In object detection pipelines, the neck component—commonly instantiated as a Feature Pyramid Network (FPN) [25], Path Aggregation Network (PAN) [28], or other multi-scale fusion structure—plays a pivotal role in aggregating features from different spatial resolutions. This stage serves as a crucial bridge between the backbone and the detection head, enriching feature maps with both low- and high-level information needed for accurate localization and classification. Knowledge distillation at this level, therefore, offers a unique opportunity to improve multi-scale representation in student models by transferring contextually enhanced features from a more capable teacher network.

A central axis in neck-level distillation is spatial importance modulation. Early approaches assumed that all spatial locations in a feature map are equally informative, which leads to diluted or misaligned supervision. To address this, techniques such as Feature Richness Score (FRS)-based distillation [29] introduce semantic-aware weighting schemes that prioritize regions contributing more significantly to detection outcomes (Fig. 3.a). These scores, often computed as the maximum predicted class probability at each spatial location, are used to generate soft attention masks. These masks then weight feature loss, guiding the student to focus on semantically valuable areas. This approach enhances spatial precision, particularly for object-centric regions, but its efficacy can be limited when the student and teacher use heterogeneous architectures, due to differing feature dimensionalities, spatial resolutions, or semantic alignment across corresponding locations. In such cases, direct feature matching may become unreliable, as similar spatial positions may encode different information or have incompatible structural representations.

To accommodate architectural mismatch and reduce overreliance on absolute feature values, a separate family of techniques leverages structure-aware statistical alignment, notably using correlation-based metrics. For example, methods like Pearson Correlation Knowledge Distillation (PKD) [30] normalize feature maps to zero mean and unit variance before computing loss, focusing on the relational structure rather than absolute activations (Fig. 3.b). By maximizing correlation between normalized features rather than minimizing raw value differences, these techniques are more robust to outliers and dominant channels. Spatial interpolation or resizing mechanisms are commonly employed to ensure compatibility between teacher and student feature resolutions. Extensions of this idea use perceptual similarity measures, such as the Structural Similarity Index Measure (SSIM) or cosine similarity

TABLE I

BACKBONE LEVEL KNOWLEDGE DISTILLATION

| Paper (year) | Modification in Backbone | Key Strategy |
|---|---|---|
| GAN-KD (2019) [26] | Feature map mimicry using GAN | Two-stage GAN: discriminator distinguishes real/fake features |
| FGFI (2019)[20] | Object-focused regions in feature maps | Fine-grained imitation with IoU-based region masks |
| MG-KD (2019)[31] | Global and local feature emphasis | Mask-guided loss focusing on spatially important regions |
| MCNET-KD (2020)[32] | Pixel-wise & structural distillation | Heatmap matching + pairwise cosine similarity of features |
| KDGAN (2020)[33] | Multi-layer GAN feature distillation | Separate discriminators for each mimicked feature layer |
| MI-KD (2021)[21] | Multi-stage supervision of backbone | Mutual information + MSE for multi-scale features |
| DeFEAT (2021)[34] | Object/background decoupling | Region-aware MSE and KL losses with weighted categories |
| AGD-NLD (2021)[27] | Attention and relational dependency modeling | Attention maps + non-local features with guided loss |
| ReviewKD (2021)[19] | Multi-layer fusion via ABF module | Residual fusion + Hierarchical Context Loss (HCL) |
| MLKD (2022)[35] | Multi-level backbone features | Pyramid pooling + regression loss for multi-resolution distillation |
| FBKD (2022)[36] | Six-stage SSD feature distillation with GAN | L2 loss + adversarial training with adaptation layers |
| GCP-SAD (2022)[37] | Multi-scale attention distillation | Attention alignment + soft logits + box regression |
| DRKD (2023)[38] | Structural and semantic reasoning in backbone | Pixel-wise, instance-wise, and relational distillation with graph reasoning |
| SKD-EFF (2023)[39] | Attention and non-local feature distillation | Adapted attention + relational modules with combined loss |
| CUD (2024)[40] | Feature attention with feedback loop | Triple Parallel Distillation + HRAD using spatial and channel attention |
| RH-KD (2024)[41] | Cross-domain pseudo-label refinement | EMA-updated teacher ensemble with pseudo-label fusion |
| DMKD (2024)[42] | Dual attention-masked feature reconstruction | Static teacher, MSE loss on masked features |
| LMFI (2024)[18] | Logits + mask-guided feature imitation | Pearson Correlation + region-focused regression loss |
| LD-COD (2024)[43] | Shared backbone, distillation in upper layers | Continual learning via latent representations from shared backbone |

[44, 45], further enriching the taxonomic class of similarity-preserving neck distillation.

Beyond spatial and statistical refinements, other approaches emphasize semantic and instance-aware feature transfer, often relying on global prototypes and dynamic weighting schemes. Notable among these is the use of a Prototype Generation Module (PGM) and a Robust Distillation Module (RDM) as proposed in [46]. These methods extract semantic basis vectors known as prototypes that span the class-level semantic subspace. Instance-level features from both teacher and student are projected into this prototype space, and the student is trained to minimize the discrepancy in these projections (Fig. 3.c). To account for varying reliability among instances, a dynamic reliability score is computed, modulating the strength of the distillation signal based on how well-aligned a student instance is with its teacher counterpart. This results in a multi-faceted supervision strategy that combines global semantic structure with local instance fidelity. Unlike earlier methods that treat all spatial locations or features uniformly [15], including more recent neck-level KD strategies such as HEAD [47], CSKD [48], or MSSD [49], which apply global or pixelwise losses without semantic weighting, prototype-based distillation introduces a class-semantic subspace and instance-level modulation. This selective and structured transfer makes it significantly more effective in complex object detection tasks where spatial and semantic variability is high. prototype-based distillation enables more informed and selective knowledge transfer, improving generalization especially in class-imbalanced scenarios.

Another growing area within this taxonomy involves attention-modulated and feedback-based distillation [40, 50], which incorporate channel and spatial attention layers to adaptively reweight feature maps. These methods are often implemented in a looped feedback structure, where the student receives refined supervision based on its current performance, enabling progressive alignment with the teacher. This subcategory integrates principles of self-correction, dynamic loss reweighting, and cross-level interaction between features and outputs.

In sum, neck-level knowledge distillation comprises a broad and evolving taxonomy encompassing spatial saliency-driven approaches, statistical and perceptual alignment, prototype-guided semantic transfer, and attention-based dynamic distillation. Each category contributes uniquely to enhancing multi-scale feature understanding in student detectors.

Taken together, these approaches reflect a broader evolution in neck-level distillation. The table II synthesizes key trends. Most methods introduce specialized mechanisms at the neck level—ranging from attention-guided spatial masks [29, 50] and structural similarity measures [44], to semantic abstraction via prototypes [46].

TABLE II

NECK LEVEL KNOWLEDGE DISTILATION

| Paper | Modification in Neck | Key Strategy |
| --- | --- | --- |
| FRS (2021)[29] | Feature alignment using Feature Richness Score (FRS) masks | Spatial masks emphasize rich regions; L2 and weighted BCE losses for feature and logit distillation |
| PKD (2022)[30] | Normalized features aligned via Pearson Correlation | PCC-based MSE loss after normalization; handles scale and feature magnitude mismatches |
| HEAD (2022)[47] | Assistant head with homogeneous architecture attached to student neck | MSE-based feature mimicry via auxiliary head plus cross-head feature sampling |
| Ida-DET (2022)[51] | Region selection using Mahalanobis distance between proposal features | Binary masks identify high-discrepancy patches; entropy-based distillation loss |
| FGD (2022)[50] | Foreground/background separation in neck using attention maps | Dual-path: focal (masks, attention) and global (GcBlock) feature distillation |
| BKD (2022)[52] | Shared Knowledge Encoder compresses features to vectors | L2 loss between local knowledge vectors + prototype-based global alignment |
| GKD (2022)[46] | Prototype and Robust Distillation Modules with reliability weighting | Class-wise prototype matching + reliability-weighted global and local losses |
| BalancedKD (2022)[44] | SSIM applied on normalized neck features | Patch-wise luminance, contrast, and structural similarity (lSSIM) to guide student |
| IBFI-GRI (2024)[53] | Feature imitation with inconsistency-based weighting and relation modeling | L2 losses modulated by divergence maps + global relation module for rich context |
| CUD (2024)[40] | Hierarchical Re-weighting Attention Distillation (HRAD) + TPD loop | Joint feature (HRAD) and response (TPD) distillation in a feedback framework |
| GRADKD (2024)[54] | Gradient-guided and box-aware multi-scale attention distillation | Important channels and spatial regions weighted by gradient and object masks |
| MSSD (2024)[49] | Self-distillation across FPN and PAN layers | Multi-scale MSE between shallow and deep features within the same model |
| UET (2024)[55] | Uncertainty-aware feature distillation via Monte Carlo dropout | Teacher features enriched with uncertainty modeling; MSE to guide student |
| CSKD (2024)[45] | Cosine Similarity-Guided Distillation (CSKD) with assistant prediction branch | Feature and logit alignment via cosine similarity + direct MSE + assistant loss |
| CAFF (2024)[56] | CAFF-enhanced feature fusion with Pyramid Pooling Loss | Multi-scale FPN distillation using SPP and L2 losses on spatially pooled features |

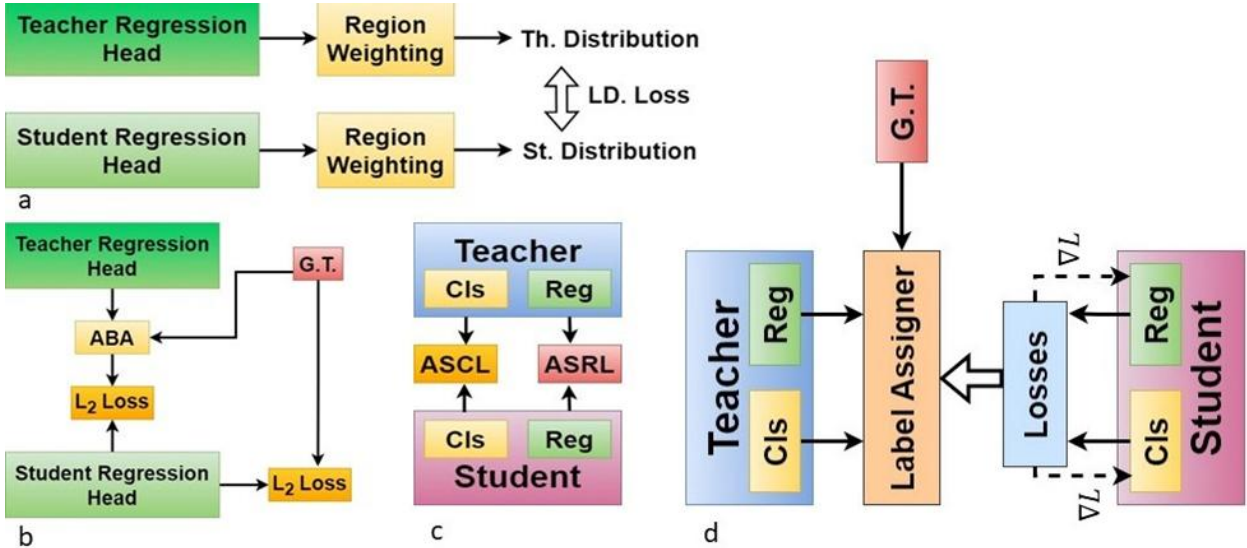

**Fig. 4.** Four methods for head level knowledge distillation. (a) Distribution-based localization distillation (LD) for regression. (b) Adaptive Boundary Approximation (ABA) for regression distillation. (c) Selective multi-head imitation for classification (ASCL) and localization (ASRL) distillation. (d) Label assignment supervision in hybrid distillation.

3) **Head-Level Distillation**

In object detection, the detection head plays a crucial role in producing task-specific outputs, including classification logits, objectness scores, and bounding box coordinates. Given that these outputs directly determine detection performance, head-level knowledge distillation (KD) is one of the most intuitive and widely adopted forms of supervision transfer. The rationale is straightforward: by guiding the student to mimic the teacher's final predictions, one can directly optimize for semantically aligned outputs. However, head-level distillation introduces several technical challenges, especially in regression scenarios, where even small discrepancies in bounding box values can drastically affect learning dynamics. Consequently, a rich taxonomy of head-level KD methods has developed, could be divided broadly into regression distillation and hybrid (multi-head) distillation frameworks.

  a) Regression Distillation

A major challenge in distilling bounding box regressors lies in the nature of the prediction space. Unlike classification outputs, which are typically discrete and amenable to soft-label transfer, bounding box predictions are continuous, sensitive to noise, and often architecture-dependent. A key innovation in this category is the formulation of bounding box predictions as probability distributions, giving rise to distribution-based regression distillation. In this paradigm, methods such as Localization Distillation (LD) [24] represent box coordinates not as single point estimates but as discrete probability distributions over coordinate bins (Fig. 4.a). This probabilistic framing allows the distillation loss formulated as KL divergence to capture the uncertainty and spatial variance present in the teacher's outputs. Importantly, this approach decouples regression supervision from hard ground-truth labels, allowing the student to benefit from the nuanced spatial priors learned by the teacher.

Despite its flexibility, distribution-based regression faces practical constraints. It often requires architectural modifications to increase output dimensionality, such as replacing direct coordinate regressors with classification-style heads that predict probability distributions over discretized bins for each bounding box coordinate. This transforms the regression task into a soft classification problem, enabling the use of KL divergence for distillation and capturing uncertainty in localization. To mitigate these concerns, a more selective approach interval-based regression distillation has emerged. Adaptive Boundary Approximation (ABA), introduced in [18], offers a representative example. Instead of modeling full distributions, ABA defines a bounded interval between the ground-truth and teacher-predicted boxes (Fig. 4.b). A piecewise regression loss penalizes student predictions only when they fall outside this reasonable interval, offering a behaviorally constrained supervision scheme. By considering the relative quality of the teacher and the ground truth, ABA introduces a mechanism to trust the teacher only when appropriate, enhancing the robustness and safety of regression transfer.

These regression-focused techniques collectively emphasize the need for fine-grained, uncertainty-aware supervision strategies. Whether through probabilistic modeling or adaptive boundary enforcement, they reflect a shift away from naive L2 regression matching and toward task-aware loss designs that account for the unique instability of localization outputs.

  b) Hybrid Distillation

Recognizing that object detection is inherently a multi-task problem, many approaches adopt hybrid head-level distillation frameworks that simultaneously supervise classification logits, objectness scores, and bounding box regressors. These methods aim to improve student performance through holistic imitation of the teacher's output space, often introducing architectural modifications or dynamic filters to regulate supervision flow.

One of the most representative frameworks is the selective multi-head imitation strategy [57], which separately distills classification (ASCL) and localization (ASRL) outputs (Fig. 4.c), while incorporating safeguards such as confidence weighting or IoU-based filtering. This ensures that the student is not forced to learn from unreliable or overconfident teacher predictions. Some methods employ Gaussian-masked feature representations or mask the logits to emphasize spatial consistency. These frameworks are designed to maintain semantic alignment without introducing noise from uncertain regions, making them particularly effective when the teacher is only partially accurate.

Gaussian masking is applied to emphasize spatial regions around high-confidence object centers by assigning higher weights near the center and gradually decreasing influence toward the periphery. This focuses the distillation loss on semantically important areas, reducing the impact of noisy background regions and improving alignment where teacher predictions are most reliable.

Another important method within hybrid distillation involves label assignment supervision. Rather than transferring final predictions directly, these methods influence the way the student generates its own training targets. Label Assignment Distillation (LAD) [58], for example, replaces heuristic-based anchor labeling with a probabilistic scheme derived from the teacher's outputs in LAD replaces the standard ground-truth–based label assignment. Instead of relying on heuristics like IoU thresholds to define positive and negative samples, LAD uses the teacher's confidence and spatial predictions to probabilistically determine label assignments, enabling more flexible and informed supervision (Fig. 4.d). Using mechanisms such as Probabilistic Anchor Assignment (PAA) [59], LAD allows the teacher to guide which proposals are labeled positive or negative, effectively distilling structural learning signals rather than fixed outputs. This form of indirect supervision is especially advantageous in anchor-based detectors like RetinaNet [60], where label assignment heuristics can limit generalization.

Hybrid methods may also include techniques such as cross-head feature reuse [61], soft heatmap alignment, or logit disentanglement, where classification scores are further separated into target and non-target components [17]. These variations acknowledge that the classification logit vector encodes both foreground confidence and background suppression, and seek to distill this information more explicitly.

Taken together, the taxonomy of head-level KD methods in object detection encompasses a spectrum from precise localization-aware distillation to holistic, task-integrated imitation strategies. These methods address not only the direct transfer of outputs but also the underlying training dynamics— how outputs are generated, which predictions are trusted, and how multiple heads interact.

Table III summarizes representative head-level knowledge distillation methods based on how they modify the detection head, and their core strategies. These methods are further categorized as either regression-level, focusing specifically on bounding box prediction, or hybrid, where multiple heads (such as classification, regression, and objectness) are jointly distilled. The table highlights how recent approaches go beyond basic logit imitation by introducing distribution-based learning, confidence-aware filtering, and label assignment guidance to improve the effectiveness of head-level distillation.

TABLE III

HEAD LEVEL KNOWLEDGE DISTILLATION

| Paper | Modification in Detection Head | Key Strategy | Distillation Type |
|---|---|---|---|
| KDHL (2017)[62] | Add Teacher-Bounded Regression Loss penalizes student BB's error exceeding teacher's by a margin | Margin-based upper bound lets the student surpass the teacher without penalty | Regression |
| D-YOLO (2018)[63] | Use objectness scores to weight cls/box losses | Confidence-weighted loss and region filtering | Hybrid |
| TAD (2020)[57] | Selectively distil from cls + reg heads with Gaussian-masked features and IoU-filtered boxes | Three-part multi-head KD with task-aware signals | Hybrid |
| LD (2021)[24] | Replace scalar boxes with probability distributions | KL divergence over predicted box coordinate distributions | Regression |
| RM-KD (2021)[64] | match teacher's ordering of positive anchors | KL over soft-ranked anchor scores | Hybrid |
| ST-KD (2021)[65] | Soft-heatmap distillation with early distribution softening and focal-loss adjustment | Soft label fusion into object center heatmaps | Hybrid |
| LAD (2021)[58] | Teacher influences anchor labeling via PAA | Distills via assignment logic instead of output targets | Hybrid |
| Center-KD (2021)[66] | Fuse dual-teacher heatmaps (size, offset) before distilling | Weighted fusion of dense attributes | Hybrid |
| EL-Lite (2022)[67] | KD using box confidence distributions | Match predicted regression with teacher's distribution | Regression |
| DKD (2022)[17] | Split logit KD into TCKD (target) + NCKD (non-target) | Separate KL terms for detailed logit matching | Hybrid |
| EffDstl (2023)[68] | Refine soft labels using confidence mining | Patch up weak teacher labels before transfer | Hybrid |
| Cross-KD (2024)[61] | Route student features through frozen teacher head, compute cross-head losses | Use teacher as transform module; gradients only affect student | Hybrid |
| YOLO-KD (2024)[69] | YOLOv5 decoupled head KD for cls, objectness, and soft-box regression | Mixed CE, MSE and KL loss; works on dense heads | Hybrid |
| CWKD (2024)[48] | Cross-weight cls loss by IoU and reg loss by class diff; add VDOs | Foreground-weighted distillation with extra supervision | Hybrid |
| LMFI (2024)[18] | Add Adaptive Boundary Approximation + logit KD + APTS filtering | Piecewise regression loss + confidence-guided sample use | Hybrid |

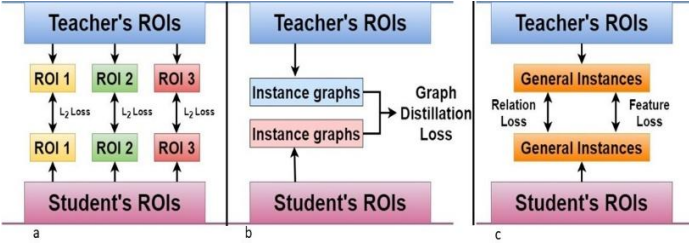

**Fig. 5.** Three methods for RPN/RoI knowledge distillation. (a) Independent feature mimicry between corresponding RoIs. (b) Graph-based distillation to learn inter-instance relations. (c) General Instance Distillation (GID) using selectively matched instances.

#### 4) RPN/RoI-Level Distillation

A critical component of the detection pipeline lies at the RPN/RoI level. This corresponds to regions of interest (RoIs), which represent candidate object locations obtained through either region proposal networks (in two-stage detectors) or dense anchors (in one-stage models). RPN/RoI-level KD offers a fine-grained strategy for aligning spatially localized, semantically meaningful features between teacher and student models. It enables more contextually grounded distillation by focusing on specific object instances, as opposed to holistic or pixelwise feature maps.

At the core of this taxonomy is the idea of **region-wise feature alignment**, where the student is trained to replicate teacher features corresponding to individual RoIs. The simplest form of this strategy involves extracting proposal-aligned features from multi-scale feature maps via RoIAlign, followed by L2 (Fig. 5.a) or cosine similarity loss between corresponding features [70]. While this basic formulation ensures spatial correspondence and preserves semantic focus, it is limited by its inability to capture relationships between regions or handle inter-instance variability effectively, because simple L2 or cosine similarity treats each RoI in isolation, ignoring the contextual dependencies and spatial interactions among multiple proposals. As a result, it fails to model how objects relate to one another—information that's critical for disambiguating overlapping or similar-looking instances in object detection.

To address these limitations, a more advanced class of methods introduces graph-based relational distillation, in which RoIs are not treated independently but rather as nodes in a structured graph encoding both individual attributes and pairwise dependencies. For instance, [22] constructs an instance graph where nodes represent proposal features and edges denote inter-instance similarity (based on cosine or Euclidean distances) (Fig. 5.b). This graph structure allows the student to learn both node-level (feature) and edge-level (relational) knowledge from the teacher. Additionally, foreground-background imbalance is explicitly addressed by incorporating both object-centric and challenging background regions into the graph, providing a more balanced and context-rich supervision signal. Such graph-based methods significantly enhance the transfer of global scene context, which is otherwise difficult to model using purely local or independent RoI features.

Despite their expressive power, graph-based methods often require strict proposal correspondence between teacher and student, which may be infeasible in detectors with divergent architectures or region proposal strategies. To relax this requirement, a separate family of methods introduces instance selection and adaptive distillation mechanisms. Notably, General Instance Distillation (GID) [71] circumvents the need for exact proposal alignment by selecting a subset of high-discrepancy instances—those for which teacher and student predictions diverge significantly. A module such as the General Instance Selection Module (GISM) ranks instances based on classification confidence differences and applies non-maximum suppression to reduce redundancy (Fig. 5.c). Distillation is then conducted across three complementary channels: (1) feature-level alignment using RoI-aligned features, (2) and relational distillation based on pairwise similarity among selected instances. This selective and multi-pronged framework enables information-efficient KD, reducing computational cost and avoiding the propagation of noisy or redundant signals. By focusing on high-discrepancy instances—where the student's predictions diverge most from the teacher's—the method targets areas with the greatest potential for informative correction. However, it selectively filters out low-confidence or uncertain predictions from both models, which are more likely to be semantically ambiguous or dominated by background, thereby minimizing the transfer of noisy supervision.

These diverse strategies collectively form a rich and emerging taxonomy for RPN/RoI-level KD. At one end of the spectrum lie approaches focusing on spatially aligned feature mimicry; at the other, we find techniques modeling structural context and semantic similarity across proposals. The key distinctions across this taxonomy include the use of explicit region correspondence (vs. selection-based filtering), the inclusion of inter-instance relational learning (vs. independent RoI supervision), and the reliance on semantic similarity metrics (vs. raw feature alignment). Furthermore, methods vary in their treatment of background regions—some explicitly incorporate hard negatives, while others filter background queries entirely to avoid noise.

The table IV summarizes key methods in the category of RPN/RoI-level knowledge distillation. These approaches focus

TABLE IV

RPN/ROI LEVEL KNOWLEDGE DISTILLATION

| Paper | Modification at RoI/Reg Level | Key Strategy |
|---|---|---|
| FM-DET (2017)[70] | RoI-based feature sampling | Mimics features from proposal regions using RPN, avoids full-map regression |
| DSIG (2021)[22] | Structured instance graph | Graph-based node and edge alignment with foreground/background handling |
| GID (2021)[71] | General instance selection | ROI selection via prediction discrepancies; feature-, relation-, and response-based distillation |
| G-DetKD (2021)[23] | Semantic fusion and contrastive learning | SGFI for adaptive semantic matching; CKD with contrastive InfoNCE loss |
| SharedKD (2024)[72] | Internal feature refinement and sharing | Cross-layer and intra-student distillation; multi-level fusion with attention |

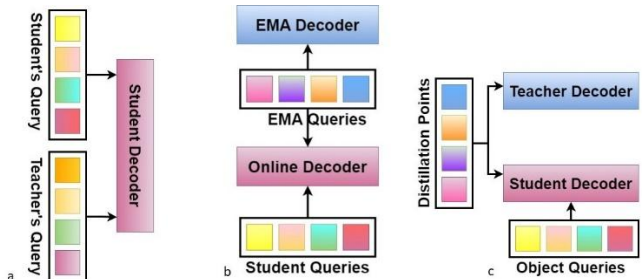

**Fig. 6.** Three methods for query-level knowledge distillation. (a) Query Prior Distillation: Injecting the teacher's queries directly into the student. (b) Temporal Query Distillation: Using an EMA teacher for online query supervision. (c) Query Disentanglement Distillation: Aligning the teacher and student with shared distillation points.

on transferring knowledge at the level of region proposals (RoIs), enabling finer spatial supervision and improved semantic alignment between teacher and student models.

### B. Transformer-Based Detectors

#### 1) Query-Level Distillation

In Transformer-based detectors, object queries are positional tokens that drive the decoding process by attending to image features through cross-attention mechanisms. Unlike grid-based detectors, these queries are learned during training and do not correspond to fixed spatial locations. This presents a core challenge: the absence of deterministic correspondence between teacher and student queries makes direct alignment ambiguous, particularly across heterogeneous architectures. Additionally, many queries tend to be redundant or background-focused, further diluting the distillation signal.

One of the solutions addresses the initialization and convergence challenges in Transformer students by transferring stable query representations from the teacher. For instance, Query Prior Assignment strategies [73] directly inject the teacher's well-trained object queries into the student model as priors (Fig. 6.a). These precondition the student's attention layers, reducing early-stage instability and guiding convergence by enforcing consistency in the query-object mapping. This class of methods can be categorized as query prior distillation, offering robust bootstrapping in low-data or low-capacity scenarios.

Another important method involves online query distillation using an Exponential Moving Average (EMA) teacher, as seen in methods like OD-DETR [74]. Instead of relying on frozen teacher weights, these approaches continuously update teacher queries to reflect the evolving feature space, then use them to supervise corresponding student queries. Auxiliary query groups, formed at intermediate decoder stages, allow deeper supervision by providing additional alignment signals beyond the final output—guiding the student not just on the end predictions but also throughout the iterative refinement process of object queries (Fig. 6.b). This helps the student learn intermediate reasoning patterns and enhances consistency across decoding layers. This strategy belongs to the temporal query distillation category, emphasizing stage-wise consistency and progressive refinement over static imitation.

A third method, query disentanglement distillation, breaks the coupling between the student's learned queries and the supervision signals. KD-DETR [75] exemplifies this with a shared set of unlearnable, architecture-agnostic distillation queries that function as anchors (Fig. 6.c). These are decoupled from detection queries and serve purely as a transfer interface. The main benefit is query repeatability and consistency, especially in heterogeneous teacher-student settings. This method is not affected by random query permutations, solving the alignment issue at its root. Together, these subclasses define a coherent taxonomy of query-level KD that addresses query initialization, alignment, redundancy, and consistency—core challenges in Transformer distillation.

To summarize the above-mentioned techniques are categorized into query prior distillation, temporal query distillation, query disentanglement, and query selection—each addressing issues like initialization stability, stage-wise alignment, architecture heterogeneity, and background noise.

#### 2) Feature-Level Distillation

Feature-level (encoder + decoder level) knowledge transfer in Transformer-based detectors differs substantially from CNN-based distillation due to the intrinsic structure of Transformer features. Unlike CNNs, which produce dense spatial grids with strong locality and translation equivariance, Transformer features are global, sparse, and generated via self-attention mechanisms. This discrepancy introduces multiple challenges for KD: feature maps between teacher and student may differ in dimensionality, positional semantics, and attention configuration. Moreover, the background dominance problem becomes acute, as many attention heads may be drawn to uninformative regions, leading to signal dilution.

One key solution is target-aware masking, where the student is guided to focus on foreground regions using soft attention masks derived from teacher object queries. For example, DETRDistill [73] applies these masks to encoder outputs, isolating the most salient areas for supervision (Fig. 7.a). This approach adapts classical feature distillation to the Transformer domain by acknowledging semantic sparsity and using teacher attention as a form of dynamic spatial masking. It represents the category of foreground-focused attention distillation.

Another method incorporates attention-guided region filtering. In QSKD [76], a selected set of high-quality queries generates foreground masks, which are used to gate the student's encoder outputs. An additional adapter layer is introduced to mitigate encoder depth mismatch, serving as a lightweight transformation module that aligns feature dimensionalities and representations between teacher and student—enabling effective distillation despite differences in depth or architecture (Fig. 7.b). This method exemplifies the attention-to-feature adaptation category, where cross-modal

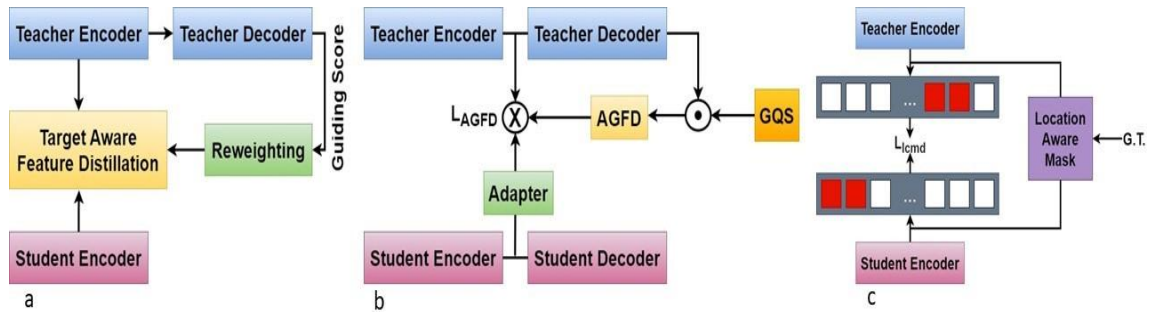

**Fig. 7.** Three feature-level knowledge distillation methods. (a) Foreground-Focused Distillation: Using teacher attention masks to focus distillation on key object areas. (b) Attention-Guided Filtering: Filtering student features using teacher-derived masks and an alignment adapter. (c) Semantic Memory Distillation: Transferring location-aware latent memory from the teacher's encoder

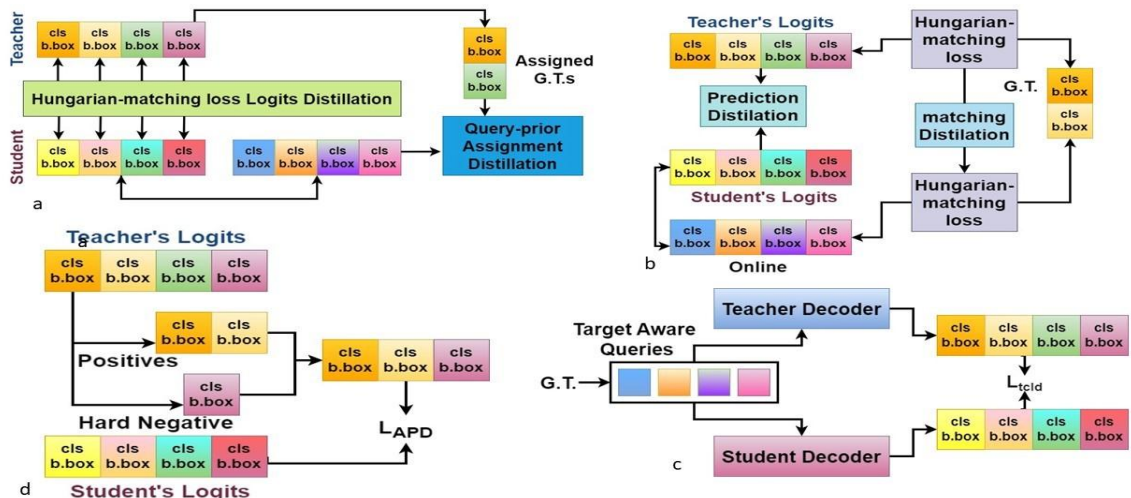

**Fig. 8.** Four strategies for logit-level knowledge distillation. (a) Matching-Based Alignment: Uses the Hungarian algorithm to pair and align student and teacher outputs. (b) Temporal Stabilization: Employs an EMA teacher to provide smoother and more stable supervision signals over time. (c) Location-Supervised Transfer: Uses ground-truth locations to guide queries, ensuring both models attend to the same regions. (d) Selective Distillation: Distills only high-value outputs (positives and hard negatives) to improve signal quality.

signals (from queries) are projected into the feature space to define the regions for distillation.

Another method incorporates attention-guided region filtering. In QSKD [76], a selected set of high-quality queries generates foreground masks, which are used to gate the student's encoder outputs. An additional adapter layer is introduced to mitigate encoder depth mismatch, serving as a lightweight transformation module that aligns feature dimensionalities and representations between teacher and student—enabling effective distillation despite differences in depth or architecture (Fig. 7.b). This method exemplifies the attention-to-feature adaptation category, where cross-modal signals (from queries) are projected into the feature space to define the regions for distillation.

Beyond masking, representation-aware feature memory transfer [77] redefines the target of feature distillation. Instead of spatial feature maps, these methods focus on encoder memory—the latent vector representations produced by self-attention modules. The memory is enriched with location-aware cues using ground-truth object annotations, ensuring spatial grounding despite the lack of explicit positional encoding in raw Transformer features (Fig. 7.c). This class, referred to as semantic memory distillation, shifts the goal from aligning pixel-like activations to transferring contextualized, object-centric embeddings.

In summary, feature-level distillation in Transformers must address fundamental structural incompatibilities between student and teacher. This taxonomy includes foreground masking, query-to-feature projection, and spatially enriched memory embedding as key strategies to transfer semantic content while minimizing irrelevant or noisy supervision.

3) **Logit-Level Distillation**

Perhaps the most conceptually challenging component of KD in Transformer-based object detectors is logit-level distillation. The output of models like DETR is a set of unordered object predictions, with no fixed spatial structure and no deterministic mapping between teacher and student outputs. Each object is matched to a query via Hungarian bipartite matching, and prediction slots may change dynamically across epochs. This set-based formulation complicates any attempt to directly align logits, a task that is relatively straightforward in grid-based detectors.

TABLE V

TRANSFORMER-BASED ARCHITECTURE DIFFERENT PARTS DISTILLATION

| Method | Distillation Type | Key Strategies | Primary Benefits |
|---|---|---|---|
| DETRDistill (2022)[73] | Query, Feature, Logit | Query Prior Assignment, Target-Aware Feature Masking, Progressive Logits Alignment | Addresses query mismatch, adapts CNN-to-Transformer feature mismatch, stabilizes training |
| OD-DETR (2024)[74] | Query, Logit | EMA Teacher Guidance, Auxiliary Query Groups, Matching Distillation | Improves query and logit stability, enhances convergence, avoids extra inference cost |
| KD-DETR (2024)[75] | Query | Decoupled Distillation Queries (Unlearnable), Architecture-Agnostic Supervision | Enables stable and repeatable supervision across heterogeneous architectures |
| QSKD (2024)[76] | Query, Feature, Logit | Group Query Selection (GQS), Attention-Guided Feature Distillation, Selective Logits | Reduces noise, emphasizes informative queries, integrates hard negatives |
| DLIM-Det (2024)[78] | Query | Query Position & Relation Distillation, Frozen Teacher Backbone | Aligns spatial and relational query features, promotes efficient training |
| CLoCKDistill (2025)[77] | Feature, Logit | Location-Aware Memory Embedding, Ground Truth-Guided Decoder Supervision | Enhances semantic alignment, reduces background bias, leverages spatial priors |
| SO-DETR (2025)[79] | Feature | Dual-Domain Feature Fusion (FFT + Spatial), Query IoU Selection, KD Loss on Decoder | Improves small object detection, leverages frequency-spatial integration with lightweight backbones |
| SSD+IOR KD (2025)[80] | Feature | Semantic Spectral Decomposition, Inter-Object Relationship Similarity | Transfers both semantic and relational context; sharpens attention to foreground |
| DCA (2025)[81] | Feature, Logit | Divide-and-Conquer Recognition, Semantic-Driven Queries, Duplex Classifier Fusion | Mitigates recognition forgetting, embeds PLM-based semantics, improves incremental generalization |
| MFD-KD (2025)[82] | Feature, Logit | Multi-Scale Frequency Domain FFT, Gaussian Weighting, Logit Similarity Loss | Enhances robustness to noise and multi-resolution features; works across CNN and Transformer |
| EA-DETR (2025)[83] | Feature, Logit | Modifies Hungarian loss and uses masked MSE loss | Simplifies distillation process using already existed parts |

The first class of logit KD methods addresses this with matching-based logit alignment. For example, DETRDistill [73] uses the same Hungarian matching algorithm used in detection training to pair student and teacher outputs based on object identity (Fig. 8.a). Once aligned, standard loss functions (KL divergence or MSE) can be applied to class scores and bounding box logits. This method directly leverages the set-based nature of Transformer outputs but suffers from instability in early training due to fluctuating matches, because the student's predictions are still poorly formed, leading to inconsistent pairings with the teacher's outputs during bipartite matching—making the supervision signal noisy and less reliable.

To mitigate this instability, methods such as OD-DETR [74] introduce temporal stabilization techniques. Using an EMA teacher, they provide smoother supervision signals that evolve with the student's learning. In addition to matching predictions across models, these techniques incorporate auxiliary query groups and stage-specific distillation, helping the student form stable logit representations (Fig. 8.b). This class, temporal ensemble distillation, smooths the supervision trajectory and is especially effective in online learning scenarios.

A third method redefines the supervision space through target-aware and location-conditioned distillation. CLoCKDistill [77] introduces ground-truth-informed decoder queries, ensuring that both the student and teacher attend to consistent object regions. This sidesteps the matching problem by anchoring attention on annotated locations (Fig. 8.c). As a result, logit alignment becomes spatially grounded, improving both semantic consistency and robustness to background noise. This class can be labeled location-supervised logit transfer.

Lastly, selective distillation method QSKD [76] implement filtered logit alignment, where only logits corresponding to high-value queries—those with strong GIoU or classification scores—are distilled. By removing redundant or low-confidence outputs, this signal-pruned distillation approach reduces overfitting to noisy teacher predictions and accelerates convergence (Fig. 8.d). This is particularly useful when the student has a smaller query budget or reduced capacity.

Collectively, logit-level KD in Transformers involves strategies that either impose structured correspondence (e.g., matching), condition supervision on prior knowledge (e.g., location), or dynamically prune the output space. Each addresses the core challenge of permutation invariance and output instability, central to Transformer-based object detection.

Table V outlines the rest of key transformer-based distillation methods, each addressing DETR-specific challenges such as unordered queries, background noise, and unstable predictions.

## IV. EVALUATION PROTOCOLS IN KNOWLEDGE DISTILLATION FOR OBJECT DETECTION

### A. Evaluation Metric

Mean Average Precision (mAP) is the standard metric for evaluating object detection performance, capturing both

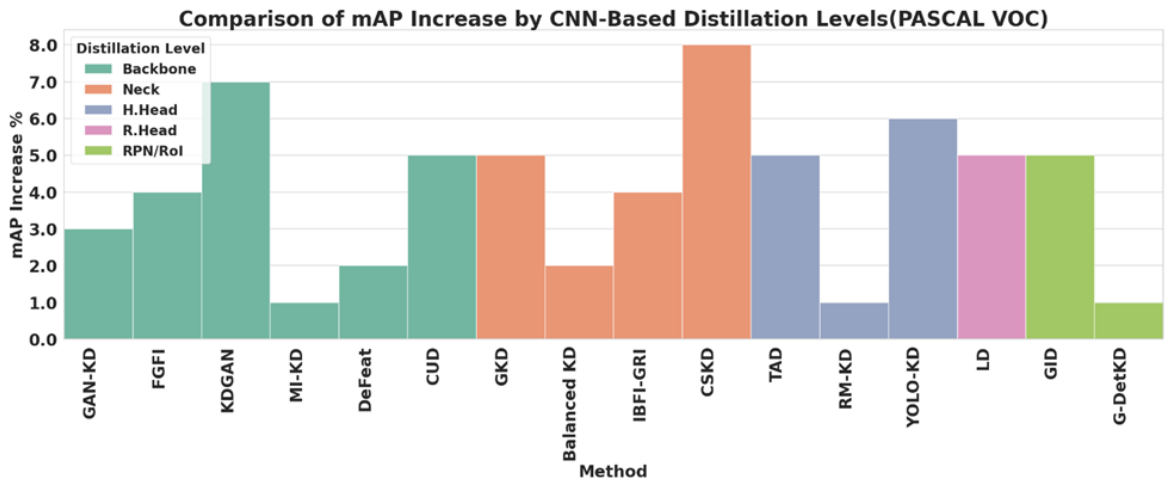

**Fig. 9.** Comparison of mAP increase by CNN-based distillation levels on the PASCAL VOC dataset.

precision and recall across various levels of localization accuracy.

mAP@0.5: Calculates the average precision at a single Intersection over Union (IoU) threshold of 0.5. This metric is commonly used in the PASCAL VOC challenge and provides a straightforward measure of detection accuracy.

*B. Datasets*

The choice of dataset greatly influences the training and evaluation of KD methods in object detection. This review focuses on two widely adopted benchmarks:

MS COCO (Common Objects in Context) [84]: A large-scale dataset with over 200,000 labeled images spanning 80 object categories. It is known for its complex scenes, multiple objects per image, and varying object sizes. COCO is a standard benchmark for evaluating modern object detectors due to its diversity and difficulty.

PASCAL VOC [85]: Contains annotated images across 20 object categories. It features relatively simpler scenes than COCO and is often used for preliminary evaluation and model comparisons. VOC is closely associated with the mAP@0.5 metric.

Note that while other metrics such as inference time or datasets like KITTI [86] and Cityscapes [87] are relevant in broader contexts, they are excluded from this review to maintain a focused and consistent evaluation framework, and due to the varying reporting standards across papers for these other metrics and datasets.

## V. COMPARATIVE ANALYSIS OF THE METHODS

*Understanding the Evaluation Metric*

All the following three comparative figures in this review use the normalized mean Average Precision gain metric, expressed as:

$$\frac{\Delta(mAP)}{\text{mAP}_{\text{baseline}}} = \frac{(\text{mAP}_{KD} - \text{mAP}_{\text{baseline}})}{\text{mAP}_{\text{baseline}}}$$

Here:

- $\text{mAP}_{KD}$ is the detection performance of the student model after applying a knowledge distillation method.
- $\text{mAP}_{\text{baseline}}$ is the detection performance of the same student model trained without any distillation.
- The resulting value quantifies the relative improvement due to KD as a percentage gain over the original performance, rather than absolute mAP increase. This normalization allows fair comparison across architectures with different baseline performances.

Fig. 9 benchmarks CNN-based distillation strategies applied to PASCAL VOC using mAP@0.5. Backbone-level knowledge distillation (KD) methods show moderate but consistent gains, with hierarchical supervision and distribution aligning approaches such as KDGAN [33] and CUD [40] leading the category. These techniques leverage spatial and semantic cues to enhance low- and mid-level feature representations, which are particularly effective on the simpler spatial layouts typical of VOC. Neck-level KD techniques, such as CSKD [45], demonstrate relatively larger normalized gains, suggesting that multi-scale fusion is critical for VOC, where the limited object variance benefits from spatial modulation. At the head level, KD strategies that combine classification and regression components, like YOLO-KD [69], outperform other hybrid-head distillation methods while LD alone has good impact on performance gain comparable to hybrid method indicating the importance of regression head distillation. Finally, RPN/RoI-level KD methods, exemplified by General Instance Distillation (GID) [71] exhibit strong results due to precise supervision at the proposal level, emphasizing the importance of spatial alignment and instance-aware learning in cleaner datasets such as PASCAL VOC.

Overall, it can be seen that distillation at every level has a significant effect on the final mAP gain, with particular emphasis on the backbone and neck levels, which produce the essential features for object detection and are therefore critical in the distillation process for PASCAL VOC dataset.

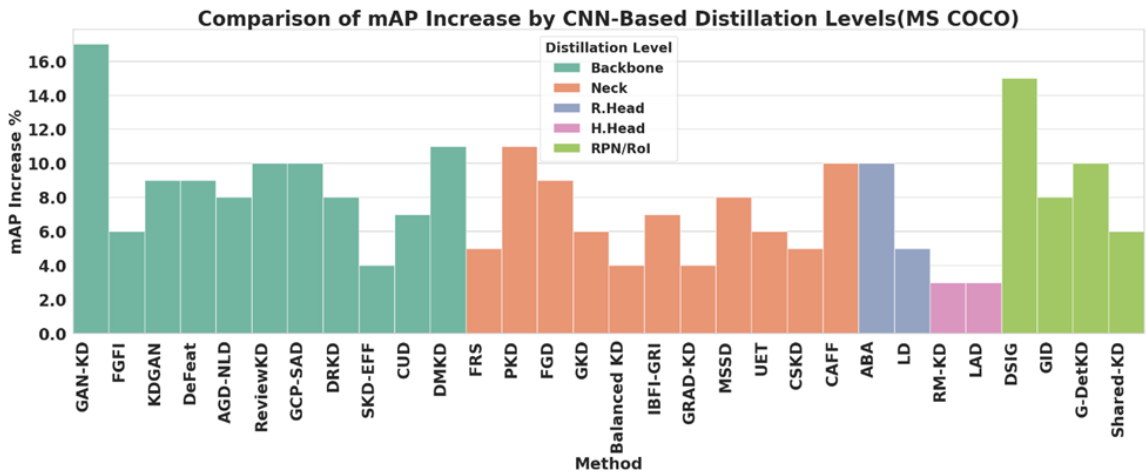

**Fig. 10.** Comparison of mAP increase by CNN-based distillation levels on the MS COCO dataset.

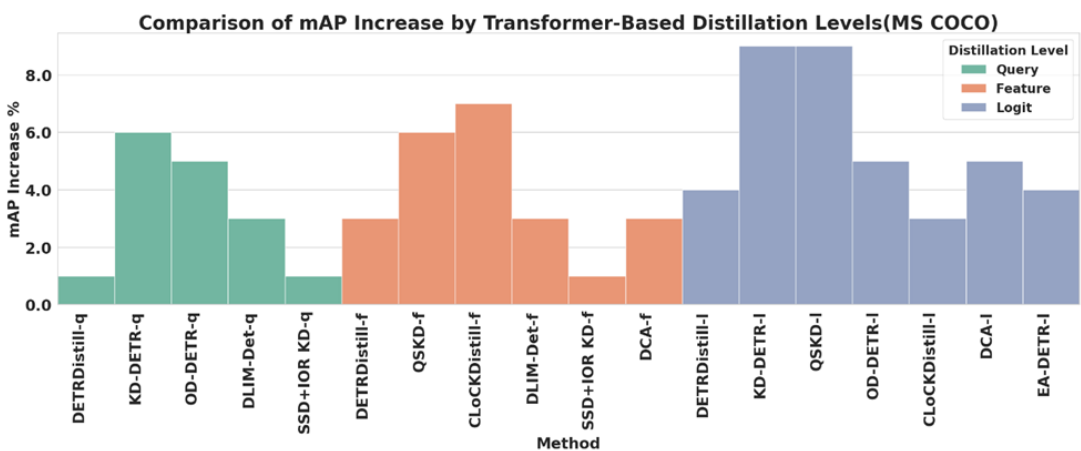

**Fig. 11.** Comparison of mAP increase by Transformer-based distillation levels on the MS COCO dataset.

MS COCO presents a more challenging benchmark, with higher object density, scale variance, and background clutter. Analysis of normalized gains in Fig. 10 reveals several patterns:

At the backbone level, GAN-based distillation methods show significant improvements in mAP similar to PASCAL VOC case, indicating that distribution-based distillation is more effective for feature-level knowledge transfer. In neck-level distillation, score-based methods like PKD [30] achieve notable gains, suggesting that transferring generalized forms of knowledge is highly effective even for large datasets like MS COCO. At the head level, hybrid distillation strategies once again lead in performance; however, regression-only methods show comparable mAP increases, highlighting that head-level distillation is essential for effective knowledge transfer.

Finally, in RPN/RoI-level distillation, methods such as DSIG [22] achieve mAP gains comparable to those at the backbone and neck levels, underscoring the importance of this level in the distillation process. Overall, as seen with the PASCAL VOC dataset, distillation at all architectural levels can significantly impact performance on MS COCO, with backbone and neck-level distillation leading the improvements. It can be concluded that, despite changes in dataset complexity, the overall comparison shows consistent performance gains at every level for both the simple PASCAL VOC and the more complex MS COCO datasets.

Fig. 11 evaluates KD methods for Transformer-based detectors, measured on COCO with mAP@0.5. The figure indicates that in transformer-based object detection (OD), Among the highest performing methods, several are logit-level, suggesting its strong potential role in enhancing the quality of knowledge transferred between teacher and student models. This is evidenced by methods like KD-DETR [75] and QSKD [76], which show substantial improvements in mAP, highlighting the effectiveness of aligning output distributions, especially in models that rely heavily on global context modeling. In contrast, both feature-level and query-level distillation exhibit moderate yet consistent performance gains.

These methods focus on transferring intermediate representations and attention mechanisms, which help the student model better capture spatial and semantic relationships. Among these, ClockDistill [77] stands out for its superior performance in feature-level distillation, likely due to its temporal alignment strategy that better preserves the progressive refinement of features. Overall, the results suggest that while feature and query-level strategies contribute meaningfully to model compression and efficiency, logit-level

Table VI

CNN-BASED KD IN OD (B.: BASE, D.: DISTILATION, B:BACKBONE, N:NECK, HH: HYBRID HEAD, RH: REGRESSION HEAD, R: RPN/RoI, Het: HETEROGENITY

| Method | Teacher → Student | B. mAP | D. mAP | Gain | D. Lvl(s) | Het |
|---|---|---|---|---|---|---|
| **GAN-KD [26]** | R101 → MobileNetV1 | 75.4 | 77.6 | +2.2 | B | No |
| **AGD-NLD [27]** | RNeXt101 → R50 (Faster R-CNN) | 38.4 | 41.5 | +3.1 | B | No |
| **Review-KD [19]** | R101 → R50 (Faster R-CNN) | 37.93 | 40.36 | +2.43 | B | No |
| **FRS [29]** | R101 → R50 (RetinaNet) | 37.4 | 39.7 | +2.3 | N (+1.5–2.0), HH (+1.0) | No |
|  | RNeXt101 → R50 (RetinaNet) | 37.4 | 40.1 | +2.7 | N + HH |  |
| **PKD [30]** | R101 → R50 (RetinaNet) | 37.4 | 40.8 | +3.4 | N | Yes |
|  | R101 → R50 (FCOS) | 39.1 | 42.8 | +3.7 |  |  |
|  | R101 → R50 (Faster R-CNN) | 38.4 | 40.5 | +2.1 |  |  |
| **GKD [46]** | R101 → R50 (Faster R-CNN) | 38.4 | 40.6 | +2.2 | N + R | No |
|  | RNeXt101 → R50 (Faster R-CNN) | 38.4 | 41.5 | +3.1 |  |  |
|  | R101 → R50 (RetinaNet) | 37.4 | 39.8 | +2.4 |  |  |
| **ABA [18]** | R101 → R50 (RetinaNet) | 33.4 | 38.1 | +4.7 | N (+0.9), RH (+3.6) | No |
|  | R101 → R50 (FCOS) | 34.3 | 38.0 | +3.7 | N + HH |  |
| **LD [24]** | R101 → R50 (RetinaNet) | 36.9 | 39.0 | +2.1 | RH | No |
|  | R101 → R50 (FCOS) | 38.6 | 40.6 | +2.0 |  |  |
| **TAD [57]** | R101 → R50 (Faster R-CNN) | 36.3 | 39.0 | +2.7 | B(+2.4), HH (+3.4) | No |
|  | R101 → R50 (RetinaNet) | 35.6 | 37.9 | +2.3 | B + HH |  |
| **LAD [58]** | R101 → R50 (PAA) | 40.4 | 41.6 | +1.2 | HH | No |
| **DSIG [22]** | R101 → R50 (Faster R-CNN) | 37.47 | 39.29 | +1.82 | R | No |
|  | R101 → R50 (FCOS) | 37.14 | 39.04 | +1.9 |  |  |
| **GID [71]** | R101 → R50 (RetinaNet) | 36.3 | 39.1 | +2.8 | R | No |
|  | R101 → R50 (Faster R-CNN) | 37.4 | 39.5 | +2.1 |  |  |
|  | R101 → R50 (FCOS) | 37.1 | 39.7 | +2.6 |  |  |

distillation remains the most impactful technique for boosting accuracy in transformer-based object detectors.

As previously noted, the majority of the reviewed methods utilize multi-stage distillation frameworks. To evaluate the effectiveness of each method individually, a subset of the most impactful approaches has been selected and presented in the following tables, with a focus on knowledge distillation for CNN-based and Transformer-based object detectors.

One of the clearest insights from the table VI is that distillation strategies focusing on the head — particularly hybrid head-level distillation that addresses both classification and localization outputs — consistently deliver the most significant performance gains. This is evident in methods like ABA [18], which achieved a +4.7 mAP gain on RetinaNet by applying logit-level mimicking for both tasks alongside feature imitation, demonstrating the advantage of transferring task-specific decision boundaries. Similarly, TAD's [57] gains from classification and regression head distillation confirm that bounding box supervision is critical. In contrast, backbone-only approaches such as GAN-KD [26] or AGD-NLD [27], though effective, offer modest improvements because they transfer generic features rather than task-tuned representations. Neck-level distillation, especially when combined with head-level targets as seen in FRS [29] and PKD [30], provides a middle ground by enabling multiscale semantic guidance. PKD [30] notably leverages correlation-based loss to robustly align multi-resolution FPN features, showing that effective multi-scale feature alignment at the neck, which directly feeds the head, is crucial to yields better generalization and learning for students.

A major shortcoming exposed by the table is the widespread lack of support for heterogeneous architecture distillation — that is, distilling knowledge across different types of detection architectures, such as from two-stage to one-stage models or from CNN-based to Transformer-based detectors. Except for PKD [30], which explicitly supports such cross-architecture transfers (e.g., Mask R-CNN with Swin Transformer to CNN-based RetinaNet), nearly all other methods are constrained to intra-architecture settings. This limitation likely stems from the reliance on spatial feature alignment and structural symmetry, which breaks down when the architectural inductive biases diverge. Consequently, there is a clear research gap in developing architecture-agnostic distillation techniques that are resilient to representation disparities between teacher and student.

The comparison also reveals a trade-off between performance gains and methodological simplicity. Lightweight approaches like LAD [58] or Review-KD [19] offer minimal improvements (around +1 to +2 mAP) but are attractive for their minimal computational burden and straightforward implementation. In contrast, complex methods like LMFI [18], PKD [30], or GID [71] integrate multiple distillation components or cross-instance modeling strategies and consistently report higher gains. However, their complexity may hinder scalability, especially in real-time or resource-constrained scenarios, suggesting that practical applicability should always be balanced against theoretical effectiveness.

Graph-based and relation-aware distillation methods, such as DSIG [22] and GID [71], stand out for their superior handling of object detection in complex scenes with dense or overlapping instances. These methods move beyond pixel-wise or region-wise mimicry by modeling instance-level interactions and context through structured graphs or selection mechanisms. This modeling of relational information allows them to capture higher-order semantics and inter-instance dependencies that conventional methods overlook. Their superior performance

TABLE VII

TRANSFORMER-BASED KD IN OD. L: LOGIT, F: FEATURE, Q: QUERY-LEVEL DISTILLATION, Het: HETEROGENITY, S.: STUDENT

| Method | Teacher → Student | B. mAP | D. mAP | D. Lvl | Het | S. Size |
|---|---|---|---|---|---|---|
| **DETRDistill[73]** | AdaMixer R101 → R50Deformable | 42.3 | 44.7 | L(43.7) ,F(43.5) ,Q(42.9) | No | ~40M |
|  | DETR R101 → R50 | 44.1 | 46.6 | - |  |  |
|  | Conditional DETR R101 → R50 | 40.7 | 42.9 | - |  |  |
| **OD-DETR[74]** | Self-distillation (Def-DETR R50) | 45.4 | 47.7 | L(46.8), F(47.3), Q(47.2) | No | 34.7M |
| **KD-DETR[75]** | DAB-DETR R50 → R18DINO | 36.2 | 41.4 | Q(41.4) | Yes | 23M (R18), |
|  | R50 → Faster R-CNN R50 | 38.4 | 40.5 | - |  | 44M (R50) |
| **QSKD[76]** | Cond. DETR R101 → R18DAB-DETR | 35.8 | 39.9 | L(37.8), F(39.4) | No | 23M (R18) |
|  | R101 → R18 | 36.2 | 41.5 | - |  |  |
| **CLoCKDistill[77]** | DINO R50 → R18DAB-DETR | 56.1 | 62.5 | F(MEM59.9) F+Mask(61.6) + L(62.5) | No | 31M (R18) |
|  | R50 → R18 | 39.8 | 45.2 | - |  |  |
| **DLIM-Det[78]** | Swin-L → Swin-T | 51.3 | 54.4 | Q(52.1) + L(QPD: 51.6) + F(QRD: 51.8) | No | 28M |
|  | Intern-L→Intern-T | 53.2 | 55.9 | - |  | ~30M |
|  | ResNet101 →ResNet50 | 48.9 | 49.8 | - |  | ~47M |

underlines the potential of incorporating structural awareness into the knowledge transfer process for object detectors.

Interestingly, a recurring trend among several methods — particularly FRS [29], PKD [30], and GKD [46]— is that distilled student models can outperform their teachers. This turns the classical KD assumption on its head, illustrating that KD can serve not just as a compression mechanism but also as a tool for optimizing and refining knowledge representations. It suggests that teachers often contain redundant or suboptimal patterns, and a well-designed distillation process can help students internalize a more distilled, noise-free version of the task-relevant knowledge, potentially even leading to generalization beyond the teacher's own capacity.

Among the knowledge distillation methods for transformer-based object detection, KD-DETR [75], QSKD [76], and CLoCKDistill [77] demonstrate the highest efficiency in terms of mAP improvement relative to student model size. These methods consistently achieve over 2 AP gain per 10 million parameters, particularly when applied to lightweight students such as ResNet-18 [88], Swin-Tiny [89], or InternImage-Tiny. In contrast, methods like DETRDistill [73] and OD-DETR [74], which target larger students (e.g., ResNet-50), provide more modest relative gains despite comparable or greater model complexity. This suggests that distillation is proportionally more beneficial for compact models, where capacity is limited and external supervision can yield sharper performance increases.

Support for heterogeneous distillation, where teacher and student detectors are built on fundamentally different backbone types (e.g., transformer-to-CNN or vice versa), is notably scarce. Among the methods evaluated, only KD-DETR [75] explicitly addresses and validates heterogeneous distillation, successfully transferring knowledge from a transformer-based DINO [90] teacher to a CNN-based Faster R-CNN student. All other methods are confined to homogeneous settings within transformer-based architectures. This limitation highlights a critical gap in the current research landscape and underscores the need for more generalizable frameworks capable of bridging architectural boundaries.

An analysis of distillation level effectiveness reveals that feature-level and query-level strategies tend to outperform pure logit-level approaches, particularly in small model scenarios. For instance, QSKD [76] and CLoCKDistill [77] leverage attention-guided features or DETR memory representations to deliver gains exceeding 5 AP, while logit-only methods like DETRDistill [73] and OD-DETR [74] typically cap around 2.5 AP improvements. Query-level guidance, especially when paired with auxiliary mechanisms like attention maps or sampling strategies, proves especially potent in boosting student model accuracy. This trend suggests that deeper representational alignment between teacher and student yields more substantial benefits than final-output mimicking alone.

Comparing different methods applied to the same teacher-student configuration further reinforces the superiority of hybrid distillation strategies. For example, when distilling from DAB-DETR with ResNet-50 to ResNet-18, CLoCKDistill [77] and QSKD [76] both slightly outperform KD-DETR [75], achieving mAP gains of 5.4 and 5.3 respectively versus 5.2. Despite KD-DETR's [75] advantage in heterogeneity support, these results suggest that methods incorporating both feature and logit signals may offer more robust improvements when architectural compatibility is not a constraint. Consequently, the most effective distillation frameworks tend to be those that unify multiple levels of guidance, especially when applied to constrained student architectures.

## VI. CHALLENGES AND FUTURE DIRECTIONS

In CNN-based object detectors such as YOLO [2], Faster R-CNN [1], and SSD [3], knowledge distillation (KD) has seen more mature development, yet it still faces several architectural and functional bottlenecks. Localization knowledge is often underrepresented in existing distillation objectives, which tend to prioritize classification logits or soft labels while providing limited guidance to regression heads. Although recent approaches like localization distribution matching [24, 64] have shown potential, they often lack consistency across various detector architectures, as techniques effective for anchor-based models like RetinaNet may not transfer well to anchor-free or Transformer-based detectors, where localization signals are represented and learned differently. Furthermore, there is an over-reliance on static teacher confidence or label assignments, which can bias the student model—especially when the teacher is inaccurate or overconfident.

An emerging concern is the use of weak or shallow teacher models without a robust framework to evaluate the reliability of the supervision signals they provide during training [49]. Looking ahead, future research in KD for CNN-based detectors is expected to explore adaptive spatial resolution targeting, where feature imitation varies based on object scale and scene context. Moreover, incorporating represent a more refined extension of current distribution-based regression distillation methods. While existing approaches like LD [24] discretize coordinates into probability bins, they typically do so independently for each axis—whereas heatmap-based cues could capture spatial correlations across both dimensions, offering richer and more holistic supervision.

Adaptive teacher weighting mechanisms—where the influence of the teacher is scaled according to confidence and consistency—could also improve resilience against noisy or unreliable guidance.

Finally, relation-aware, RPN/RoI-level KD, utilizing graph-based models or contrastive learning techniques, may enable the transfer of higher-order spatial dependencies, particularly beneficial in cluttered or complex environments.

In Transformer-based object detectors like DETR [4] and its derivatives, knowledge distillation is still in an early stage, primarily due to the unique challenges posed by set-based prediction and decentralized query processing. One major challenge is the instability and semantic inconsistency of learned queries. Although techniques like KD-DETR [75] and QSKD [76] attempt to align decoder outputs using shared or selected queries, the semantics of these queries often vary across tasks and training runs, reducing the generalizability of such methods.

Another limitation is the common practice of distilling knowledge only from the final decoder output, while ignoring intermediate decoder stages where critical object hypotheses are initially formed. In addition, current KD strategies tend to neglect the rich self-attention and cross-attention mechanisms that are central to the spatial reasoning capabilities of Transformers. These attention patterns are seldom leveraged in distillation objectives beyond basic spatial alignment.

Moreover, most approaches fail to exploit the global and relational reasoning encoded in the attention matrices and encoder memory of the teacher. Applying layer-wise supervision across the decoder's temporal progression could improve convergence and robustness. Similarly, attention-guided distillation, using structured supervision such as head-wise or query-wise attention maps, may better facilitate the transfer of relational and contextual knowledge.

In the domain of heterogeneous knowledge distillation (Transformer-to-CNN) [91, 92], transferring knowledge from Transformer-based teachers to CNN-based students remains an underexplored yet critical area. Vision Transformers (ViTs [93]) have demonstrated remarkable performance due to their ability to capture long-range dependencies and model global context. However, their high computational demands make them impractical for deployment in resource-constrained environments. Conversely, CNNs are more efficient and better suited for such scenarios but often lag behind in detection performance. Recent efforts in semantic segmentation have explored cross-architecture distillation, where knowledge is transferred not only from CNNs to Transformers but also in the reverse direction. Some methods support bi-directional distillation [94, 95], demonstrating that Transformer-to-CNN knowledge transfer can effectively boost the performance of lightweight CNN-based detectors without introducing significant computational overhead.

Crucially, investigating KD at the architectural level—by analyzing how different network types represent and propagate information—can help identify shared or compatible knowledge representations between Transformer and CNN-based detectors. This could involve developing intermediate 'translator' modules to align heterogeneous feature structures, identifying a common latent space for knowledge transfer, or designing loss functions that are invariant to architectural differences in spatial layout, attention mechanisms, or feature semantics. Understanding these underlying knowledge similarities enables more effective knowledge transfer and opens up new avenues for designing distillation strategies that are both architecture-aware and semantically aligned. These approaches highlight the potential of cross-architecture KD to bridge the performance-efficiency gap and promote the development of compact yet high-performing detection models for real-world applications.

VII. CONCLUSION

Knowledge Distillation (KD) has emerged as a vital strategy for building efficient object detectors capable of rivaling large models while remaining suitable for real-world deployment. This review has introduced a novel architectural taxonomy that classifies KD techniques by the component of the object detection pipeline they target—spanning CNN-based backbones, necks, heads, and RPN/RoI, as well as Transformer-based queries, features, and logits.

Through this taxonomy, we analyzed a wide spectrum of methods, identifying recurring design patterns, unique challenges, and specialized solutions at each architectural level. Comparative evaluations across benchmarks like PASCAL VOC and COCO further illustrated the effectiveness of tailored KD strategies, especially those addressing scale variance, semantic alignment, and output uncertainty.

By offering a fine-grained, architecture-aware perspective, this review aims to guide the development of next-generation KD strategies that balance accuracy, efficiency, and adaptability—paving the way for practical, high-performance object detection across a wide array of deployment scenarios.

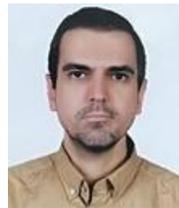

**Mahdi Golizadeh** received the M.Sc. degree in Computer Engineering. His research interests include deep learning with a focus on efficient training loops, knowledge distillation, and compact neural network design. He is particularly interested in methods that reduce computational cost while maintaining accuracy in real-world machine learning applications.

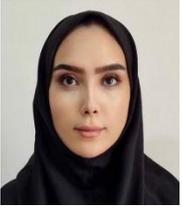
**Nassibeh Golizadeh** received the B.Sc. degree in information technology engineering and the M.Sc. degree in computer engineering from the University of Tabriz, Tabriz, Iran. She is currently pursuing the Ph.D. degree in computer engineering at Sharif University of Technology, Tehran, Iran.

As a Deep Learning Researcher at the EPCA Lab, Sharif University of Technology, her research interests include natural language processing (NLP), TinyML, AI accelerators, and knowledge distillation.

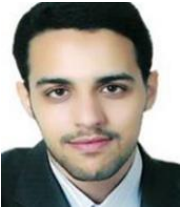
**Mohammad Ali Keyvanrad** received the B.S. degree in software engineering from the Amirkabir University of Technology, Tehran, Iran, in 2007, and the MSc and Ph.D. degrees in Artificial Intelligence from the Amirkabir University of Technology, Tehran, Iran, in 2010 and 2016, respectively. He is currently an assistant professor of artificial intelligence with the Malek Ashtar University of Technology. His research interests include pattern recognition, machine learning, especially deep learning and machine vision.

**Hossein Shirazi**, photograph and biography not available at the time of publication.